\documentclass[10pt,twocolumn,letterpaper]{article}

\usepackage[pagenumbers]{ICCV2025/iccv} 

\DeclareMathOperator*{\argmax}{arg\,max}
\DeclareMathOperator*{\argmin}{arg\,min}

\definecolor{iccvblue}{rgb}{0.21,0.49,0.74}
\usepackage[pagebackref,breaklinks,colorlinks,allcolors=iccvblue]{hyperref}
\usepackage{afterpage}
\usepackage{graphicx}
\usepackage{booktabs}
\usepackage{multirow}
\usepackage{rotating}
\usepackage{amssymb}
\usepackage{pifont}
\usepackage{pdfcomment}
\usepackage{float}
\usepackage{placeins}
\usepackage{stfloats}
\def\paperID{4618} 
\def\confName{ICCV}
\def\confYear{2025}

\title{Helping CLIP See Both the Forest and the Trees: A Decomposition and Description Approach}

\author{Leyan Xue\thanks{Equal contribution.} \\
Tianjin University \\
\and
Zongbo Han\footnotemark[1] \\
Tianjin University \\
\and
Guangyu Wang \\
Beijing University of Posts and Telecommunications \\
\and
Qinghua Hu \\
Tianjin University \\
\and
Mingyue Cheng \\
University of Science and Technology of China \\
\and
Changqing Zhang\thanks{Corresponding author.} \\
Tianjin University
}

\begin{document}
\maketitle
\begin{abstract}
Vision-Language Models (VLMs) like CLIP achieve cross-modal semantic alignment through contrastive learning, exhibiting robust zero-shot generalization. Traditional prompt engineering, however, predominantly relies on coarse-grained category labels, neglecting fine-grained local semantics. Existing approaches assume that VLMs inherently recognize localized visual details and attempt to enhance classification by augmenting text prompts with attribute descriptors generated by large language models. However, our systematic experiments reveal critical limitations: CLIP’s strong bias toward global image patterns hinders its ability to process localized visual descriptors. To address this fundamental constraint, we propose a simple, effective, and plug-and-play solution that enables CLIP to ``See Both the Forest and the Trees." Specifically, we employ stochastic multi-crop augmentation to activate CLIP’s latent capacity for localized feature analysis. By cropping only partial regions, the approach effectively constrains the model’s receptive field and recalibrates its attention mechanism, thereby mitigating its inherent bias. We evaluate the proposed method under zero-shot, few-shot, and test-time adaptation settings, and extensive experiments demonstrate that D\&D achieves promising performance.
\end{abstract} 

\begin{figure*}[htbp]
\centering
\includegraphics[width=1\textwidth]{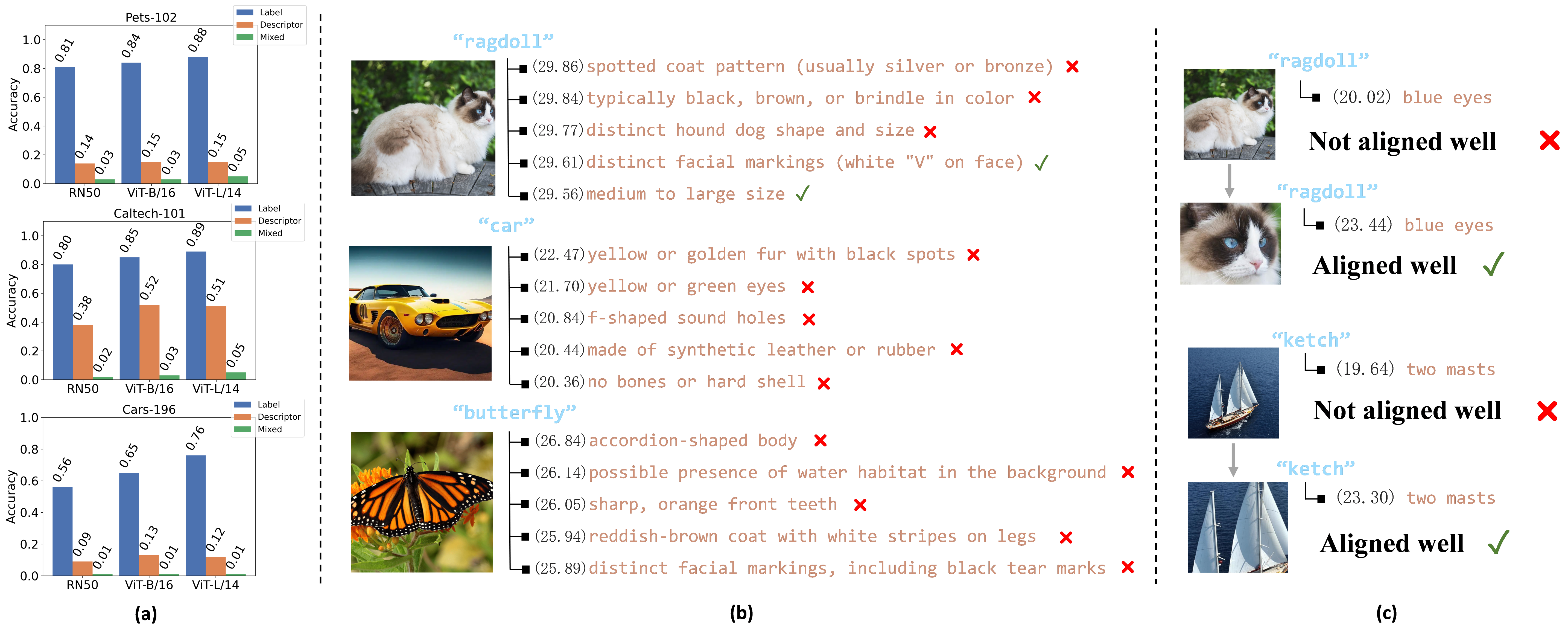} 
\caption{Motivation of the proposed method. (a) We provide a statistical analysis of the accuracy for three types of prompts. (b) We present examples of the top-5 accuracy for Hybrid Prompts, and observe that a significant proportion of these prompts do not accurately align with the corresponding labels. (c) The perception of localized descriptions by CLIP can be activated. We can use random cropping and other means to enable CLIP to align with localized features.}
\label{fig:moti}
\end{figure*}

\section{Introduction}
\label{sec:intro}
\textcolor{black}{Vision-Language Models (VLMs) like CLIP \citep{radford2021learning} achieve cross-modal alignment of visual and textual representations through contrastive learning, demonstrating remarkable zero-shot generalization. Their core mechanism relies on semantic matching between images and textual prompts (e.g., ``a photo of a {label}"). However, traditional prompt engineering often only relies on category labels (e.g., ``dog"), failing to leverage fine-grained semantic details. To address this limitation, recent studies propose enhancing prompts by integrating large language models (LLMs) \citep{touvron2023llama,brown2020language,pratt2023does,menonvisual} to generate category-specific attribute descriptors. For instance, LLMs might produce localized features like ``fluffy tails" or ``blue irises", expanding sparse labels into visually rich descriptions (e.g., ``a border collie with fluffy tails and blue irises"). This approach assumes that VLMs trained with contrastive learning naturally have the ability to recognize localized semantic details. By incorporating enhanced textual cues, this capability is expected to lead to more accurate classification.}

\textcolor{black}{These approaches share a similar assumption that \textit{VLMs inherently can recognize localized details in images (e.g., object attributes or textures), and that this capability can be harnessed through enriched text prompts (e.g., augmented with category-specific descriptors) to enhance classification accuracy.} However, our empirical analysis reveals that CLIP demonstrates limited perception of fine-grained local details in visual inputs. Specifically, to rigorously analyze this assumption, we introduce an experiment that disentangles the roles of class labels and local descriptors by constructing three distinct prompt types: (1) Label-only (e.g., ``a photo of a bird''),  (2) Descriptor-only (e.g., ``a red beak and striped wings''), and (3) Hybrid prompts merging labels with atypical or conflicting descriptors (e.g., ``a bird with blue wheels'').  
Evaluations across 3 datasets and 3 model architectures, along with the experimental results, are shown in Fig.~\ref{fig:moti}.}

\textcolor{black}{From the experimental results, we can have the following two observations. Firstly, CLIP achieves an average zero-shot accuracy of 78.24\% with label-only prompts, but its performance drops sharply to 24.31\% with descriptor-only prompts, indicating that CLIP's classification predominantly depends on label prompts while exhibiting limited perception of localized descriptors. Secondly, label-descriptor hybrid prompts yield extremely low accuracy (2.28\%) when evaluated under a strict criterion: a prediction is correct only if the model identifies both the label and its corresponding descriptor. This experiment is designed to assess CLIP's ability to integrate both global category information (labels) and fine-grained details (descriptors) for accurate classification. The result suggests that CLIP may not actually be capable of recognizing localized descriptors, instead treating textual descriptions as mere contextual modifiers.}

\textcolor{black}{To further analyze this phenomenon, we conducted a deeper experiment from the perspective of similarity. Specifically, we performed a similarity-level analysis comparing CLIP’s responses to label and descriptor perturbations. For an image of a bird, we computed the similarity difference between a matched mixed prompt (e.g., ``bird with a red beak") and a mismatched mixed prompt (e.g., ``bird with blue wheels"), finding an average drop of 0.23. In contrast, the similarity difference between label-matched and unmatched prompts (e.g., ``bird" vs. ``car") was 5.91, demonstrating that CLIP’s predictions are far more sensitive to label perturbations than to descriptor changes. In Fig.~1(b), we present several examples along with their top-5 highest similarity score hybrid prompts, showing that despite the high similarity scores, their fine-grained semantic descriptors  usually do not correspond well to the images.}

\textcolor{black}{The above experimental results indicate that CLIP is more sensitive to global information than to local features. This key observation inspired us to enhance CLIP’s ability to capture local details by transforming localized visual elements into global representations. Specifically, we propose a simple yet effective strategy: random cropping. As shown in \textcolor{black}{Fig.~1(c)}, this simple method enables the CLIP to explicitly align regional visual patterns with their corresponding textual prompts. By cropping only partial regions, the approach effectively constrains the model’s receptive field and recalibrates its attention mechanism, thereby mitigating its inherent bias. }

\textcolor{black}{Building on this insight, we introduce a simple method to enable CLIP to capture both global and local features by decomposing images and describing classes via prompts (D\&D). We evaluate the proposed method under zero-shot, few-shot, and test-time adaptation settings, and extensive experiments demonstrate that D\&D achieves promising performance. Our contributions are as follows: (1) a extensive evaluation revealing CLIP’s gap in local descriptor perception, (2) a plug-and-play adaptation method helping CLIP ``See Both the Forest and the Trees'', (3) extensive experiments under multiple settings validate the effectiveness of the proposed method.}


\section{Related Works}
\label{sec:related}

\subsection{Vision-Language Models}
Vision-Language Models (VLMs) have become a cornerstone in bridging the gap between vision and language. By aligning the visual and linguistic representations in a shared feature space, VLMs are capable of performing a wide range of multimodal tasks. Notable models like CLIP \citep{radford2021learning} and ALIGN \citep{jia2021scaling} employ contrastive learning to achieve this alignment, producing highly generalizable representations. VLMs have found applications in various domains, including image classification \citep{radford2021learning}, object detection \citep{guopen} and semantic segmentation \citep{lilanguage}. Additionally, VLMs have shown great promise in open-world tasks such as open-world
segmentation \citep{xu2022simple,ding2022decoupling} and open-world detection \citep{joseph2021towards,gupta2022ow}, demonstrating their versatility and scalability in real-world scenarios. Recent research has revealed several limitations of CLIP \citep{roth2023waffling,mao2023doubly,yuksekgonuland}. In our study, we identified that CLIP exhibits poor sensitivity to local features. To address this issue, we propose the use of random cropping as a mitigation strategy.

\subsection{Adaptation of VLMs}
Contrastive Language-Image Pre-Training (CLIP) \citep{radford2021learning} is a prominent vision-language model (VLM) that aligns images and texts in a shared feature space using contrastive learning.  Its performance can be further improved through various adaptation techniques. For instance, integrating large language models (LLMs) to dynamically generate task-specific prompts has been shown to enhance CLIP's capabilities by providing more context-aware representations\citep{touvron2023llama,brown2020language,pratt2023does,menonvisual}. Few-shot learning techniques like CoOp\citep{zhou2022learning} and Tip-Adapter \citep{zhang2021tip} fine-tune CLIP for specific tasks, optimizing textual prompts or adding lightweight adapters \citep{gao2024clip,zhou2022conditional} to improve accuracy with minimal data. Additionally, Test-Time Adaptation (TTA) methods \citep{abdul2023align,karmanov2024efficient} allow CLIP to adapt during testing, making it more robust across different domains without further training. In our study, we chose to make improvements based on Tip-Adapter \citep{zhang2021tip} and TDA \citep{karmanov2024efficient}.

\subsection{Optimal Transport}
Originating from Gaspard Monge’s \citep{monge1781memoire} formulation of resource redistribution, Optimal Transport (OT) formalizes the problem of finding an optimal coupling between two distributions that minimizes a transportation cost defined over a ground metric space. OT has emerged as a foundational tool across diverse fields of machine learning, computer vision, and data science, driven by its ability to geometrically align structured data \citep{peyre2019computational,zhang2020deepemd}. Leveraging its strong distribution matching capabilities, OT has found applications across various domains, including generative modeling \citep{arjovsky2017wasserstein,choi2024scalable,zhaoneural} , domain adaptation \citep{courty2016optimal,fatras2021unbalanced, xu2020reliable}. Recent work has applied OT to diverse fields, including point cloud registration \citep{shen2021accurate}, unsupervised video action segmentation \citep{xu2024temporally}, cross-subject brain data alignment \citep{thual2022aligning}, and single-cell/spatial omics data analysis \citep{bunne2024optimal}. 
The Earth Mover's Distance (EMD) is a key metric derived from OT that quantifies the dissimilarity between two distributions by measuring the minimum cost required to transform one distribution into another~\cite{rubner2000earth}. In our study, we use the EMD distance from OT (Optimal Transport) to calculate the similarity between image modality and text modality sets, thereby achieving classification.

\section{Preliminary}
\textcolor{black}{In this section, we first introduce how to perform zero-shot classification using CLIP, then present methods to enhance its performance through fine-grained semantic descriptors. Finally, we conduct detailed experimental analyses to reveal potential problems in CLIP's visual perception capabilities regarding fine-grained semantic understanding.}
\subsection{Revisiting the Description-Enhanced CLIP}
\label{sec:3.1}
\textcolor{black}{CLIP \cite{radford2021learning} is a vision-language model that aligns images and text in a shared embedding space via contrastive learning, enabling zero-shot classification by matching image embeddings with text prompts.As shown in Fig.~\ref{fig:method1}(a), CLIP enables zero-shot classification by comparing the similarity between a test image and text descriptions (a.k.a., prompts) of specified candidate classes set. Formally, to identify the label of an image \( x \), CLIP constructs a set of classification prompts \(\{\text{Prompt}_c\}_{c=1}^C\) for the \(C\) candidate classes. \( \text{Prompt}_c \) is a sentence corresponding to class \( c \), such as `a photo of \{class\}', where the \{class\} token is replaced with the specific name of category \( c \) (e.g., `cat', `car'). The predicted label is then obtained using the following equation:
\begin{equation}
\argmax_{c \in C} \phi(\text{Prompt}_c, x),    
\end{equation}
where \(\phi(\text{Prompt}_c, x)\) is the function that calculates the cosine similarity between the \( \text{Prompt}_c \) and the image \( x \).}

\textcolor{black}{However, the above prompt often relies on category labels and fails to leverage fine-grained semantic  details about the classes. To address this limitation, recent advancements have leveraged the synergistic integration of VLMs and LLMs.  A representative method \citep{menonvisual} enhances original prompts by constructing multiple fine-grained semantic descriptors for each class. Specifically, The fine-grained semantic descriptors set $D_{c}$ for class $c$ can be obtained using large language models, such as GPT-3 \citep{brown2020language}, by asking for useful features to distinguish the specific class $c$. Then the refined prompt template follows a standardized formulation: `a photo of \{class\} which (is/has/etc) \{descriptor\}', where the descriptor is replaced with a content $d$ from the descriptor set $D_c$ for class $c$. Then, as illustrated in Fig.~\ref{fig:method1}(b), the predictive label can be  obtained as follows:
\begin{equation}
\argmax_{c\in C} \frac{1}{|D_c|} \sum_{d\in{D_c}} \phi([\text{Prompt}_c, d], x),
\end{equation}
where $[\text{Prompt}_c, d]$ represents combining the original zero-shot $\text{Prompt}_c$ with the specific descriptor $d$. Our specific approach to generating descriptions is detailed in the appendix \ref{appendix:build}.}
\subsection{Analysis of CLIP's insufficient perception}
\label{sec:3.2}
\textcolor{black}{As discussed in Sec.~\ref{sec:3.1}, existing methods inherently assume that CLIP naturally aligns localized visual features with text prompts containing fine-grained semantic details. This analysis aims to critically examine this assumption. To this end, we systematically compare three types of prompts: label-only prompts (e.g., ``a photo of a bird''), descriptor-only prompts (e.g., ``a red beak and striped wings''), and hybrid prompts, which integrate labels with descriptors (e.g., ``a bird with beak and striped wings'').}

\textcolor{black}{Our findings, as presented in Fig.~\ref{fig:moti}(a), reveal its limitations in integrating localized semantic descriptors. Specifically, taking the Pets-102 dataset as an example, when faced with label-only prompts, CLIP achieves a commendable zero-shot accuracy of 88\%, indicating its robust capability to leverage global semantic information for classification tasks. However, when the model is presented with descriptor-only prompts, its accuracy significantly drops to 15\%. This stark contrast reveals CLIP's challenge in interpreting and integrating fine-grained, localized semantic details from the descriptors. Contrary to our expectations, the CLIP model does not effectively utilize these fine-grained semantic descriptors, such as ``fluffy tails" or ``blue eyes," for accurate classification.}

Moreover, we designed an experiment where a prediction is only considered correct if the model identifies both the label and its corresponding descriptor. This experiment is designed to assess CLIP's ability to integrate both global category information (labels) and fine-grained details (descriptors) for accurate classification. Under this setting, the accuracy plummets to an extremely low level, with the highest being 5\% for the Pets dataset and the lowest being 1\% for the Cars dataset. This indicates that CLIP struggles to effectively integrate global category information (labels) with fine-grained details (descriptors),  highlighting a potential limitation in its ability to focus on localized descriptors rather than broader category information provided by labels. This highlights a key difference between human reasoning and CLIP's approach. Humans typically rely on fine-grained details to make accurate classifications, whereas CLIP primarily depends on the overall label rather than integrating specific descriptors.\\
\begin{table}[htbp]
    \centering
  \caption{Similarity analysis of hybrid prompt.}
  \footnotesize
    \begin{tabular*}{\linewidth}{c|cccc}
    \toprule
    \textbf{Dataset} & \textbf{Intra} & \textbf{Cross} & \textbf{$\Delta$ Similarity} & \textbf{$\Delta$ Label-Similarity} \\
    \midrule
    \textbf{Caltech} & 23.87 & 23.08 & 0.79    & 9.04 \\
    \textbf{Pets} & 26.50  & 26.41 & 0.09     & 10.18 \\
    \textbf{AirCraft} & 23.72 & 23.70  & 0.02  & 4.16 \\
    \textbf{EuroSAT} & 18.6  & 18.52 & 0.08   & 0.99 \\
    \textbf{DTD} & 21.86 & 21.34 & 0.52   & 3.40 \\
    \textbf{Flowers} & 26.28 & 26.38 & -0.10 & 7.69 \\
    \bottomrule
    \end{tabular*}%
  \label{tab:logits}%
\end{table}%

\textcolor{black}{We conducted a similarity analysis to compare how CLIP responds to perturbations in labels and descriptors. We divided hybrid prompts into two categories. The first category consists of intra-label prompts that use descriptors matching the label, for example, ``a bird with a red beak," ensuring a clear semantic connection. The second category includes cross-label prompts that mix or even contradict the label, such as ``a bird with blue wheels." As shown in Tab.~\ref{tab:logits}, }
\textcolor{black}{we have the following observations: (1) We calculated the average difference in similarity scores by subtracting the similarity of incorrect labels from that of the correct label. For example, if ``bird'' is the correct label, we compared its similarity score to that of an incorrect label like ``car'' or ``dog''. The average difference was 5.91, as shown in the $\Delta$Label-Similarity column of Tab.~\ref{tab:logits}. This indicates that CLIP is highly sensitive to changes in global labels. (2) We compared similarity scores for intra-label prompts (e.g., ``a bird with a red beak'') and cross-label prompts (e.g., ``a bird with blue wheel''). As shown in the $\Delta$ Similarity column of Tab.~\ref{tab:logits}, the average difference was only 0.23, which suggests that descriptors have a relatively minor impact on the similarity scores compared to global labels. This indicates that CLIP prioritizes global labels over fine-grained descriptors.}

In summary, our analysis reveals that CLIP exhibits a clear preference for global label semantics over fine-grained descriptors. This tendency is evident in both the significant drop in accuracy when using descriptor-only prompts and the minimal impact of descriptors on similarity scores compared to global labels. This behavior suggests potential limitations in CLIP's ability to effectively integrate localized semantic details into its reasoning process. Such limitations may hinder its performance on tasks that require precise understanding of fine-grained visual attributes, especially when these attributes are crucial for accurate classification or reasoning.

\section{Method}
\textcolor{black}{In this section, we present our proposed method, D\&D, which leverages fine-grained semantic details by decomposing images through cropping to focus on localized regions and employing LLMs to generate detailed descriptions for the prompts. The proposed method is a plug-and-play approach. We first demonstrate the proposed method in the zero-shot setting and then apply it to few-shot and test time adaptation problems.}
\subsection{Zero-shot Classification}
\begin{figure*}[htbp]
\centering
\includegraphics[width=1\textwidth]{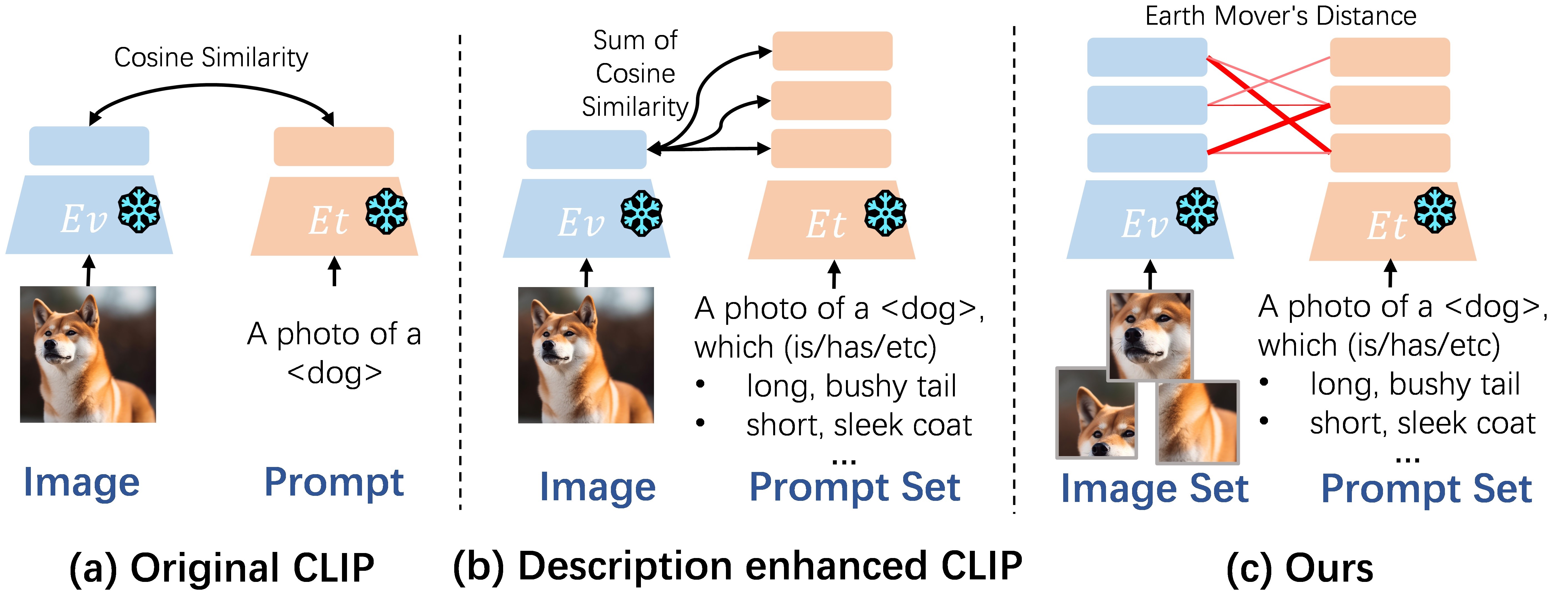} 
\caption{Overview of our proposed method under zero-shot scenario. (a) Original CLIP utilizes cosine similarity between the image embedding and text embedding to match images with text prompts. (b) Description Enhanced CLIP enhances the original CLIP by considering the sum of cosine similarities between the image embedding and multiple text embeddings derived from a set of descriptive prompts. (c) Ours: we introduce D\&D, a novel approach using Earth Mover's Distance (EMD) to compare the set of features between the image and text embeddings, potentially offering more local matching capabilities.}
\label{fig:method1}
\end{figure*}
\textcolor{black}{
\textbf{Key motivation of the proposed method}. 
The proposed method is motivated by critical observations of CLIP's limitations in zero-shot classification scenarios. Specifically, CLIP’s inherent bias toward global image-text alignment tends to overlook discriminative local visual features, resulting in suboptimal performance when fine-grained or less prominent regions are essential for accurate classification. Moreover, as illustrated in Sec.~\ref{sec:3.2}, existing description-enhanced CLIP methods do not fully leverage local information.}


\textcolor{black}{To explore how to reawaken CLIP's potential in extracting local features, we propose randomized local cropping, an inference-time strategy that processes multiple randomly cropped regions (e.g., 10–75\% of the original image scale). As shown in Fig.~\ref{fig:moti}(c), we find that random cropping allows CLIP to focus more on local features and improve the capabilities for fine-grained semantic understanding. Inspired from this, as shown in Fig.~\ref{fig:method1}(c), we propose D\&D, which not only enhances the prompt with description but also strengthens the visual information by decomposing the image with random cropping. Specifically, we construct a set of fine-grained semantic features from an image by applying random cropping. Meanwhile, we generate a corresponding set of fine-grained semantic descriptions for the prompts using a large language model. Finally, we compute the similarity between the image and the various prompts.}

\textcolor{black}{However, calculating the similarity between these two sets is non-trivial, as the elements in the sets are not perfectly correspond to each other. To this end, we introduce the Earth Mover's Distance (EMD), which quantifies the minimal cost required to transform one distribution into another \citep{rubner2000earth}. Specifically, it can compute the minimal work required to redistribute information from visual features to text descriptions. This naturally handles set-level alignment by considering all possible soft correspondences to find the optimal latent correspondences and calculate the similarity.}

\textcolor{black}{Formally, given two sets of elements with associated feature representations, EMD computes the optimal flow between them by solving a transportation problem. Let an image $x$ be partitioned into $M$ local regions (e.g., patches) ${v_1,v_2,...,v_M}$, denoted as the set \( X = \{v_1, v_2, \dots, v_M\} \), where each \( v_m \) represents a local region of the image. Meanwhile, let $D_c={d_1,d_2,...,d_N}$ denote the $N$ fine-grained descriptors for class $c$. Then we embed all elements into a shared feature space using CLIP's encoders. The grounding cost between visual patch $v_m$ and descriptor $v_n$ is defined as:
\begin{equation}
C_{m,n} = 1 - \phi(v_m, d_n),
\end{equation}
where \(\phi(v_m, d_n)\) indicates the cosine similarity between \(v_m\) and\(d_n\). The EMD can be obtained  by seeking an optimal transport plan \( \mathbf{T} \in \mathbb{R}^{M \times N} \)  minimizing the total transportation cost:
\begin{equation}
\begin{aligned}
&\text{EMD}(x, D_c) = \min_{\mathbf{T} \geq 0} \textstyle{\sum_{m=1}^M \sum_{n=1}^N }T_{m,n}C_{m,n},\\&\text{s.t.}\textstyle{\sum_{n=1}^N T_{m,n} = \tfrac{1}{M}\forall m, \sum_{m=1}^M T_{m,n} = \tfrac{1}{N}\forall n.}
\end{aligned}
\end{equation}
Then the final prediction can be made by selecting the class with minimal EMD to the image patch set:
\begin{equation}
    \argmin_{c \in C} \text{EMD}(X, D_c).
\end{equation}}
\subsection{Few-shot Learning and Test-time Adaptation}
\begin{figure*}[htbp]
\centering
\includegraphics[width=1\textwidth]{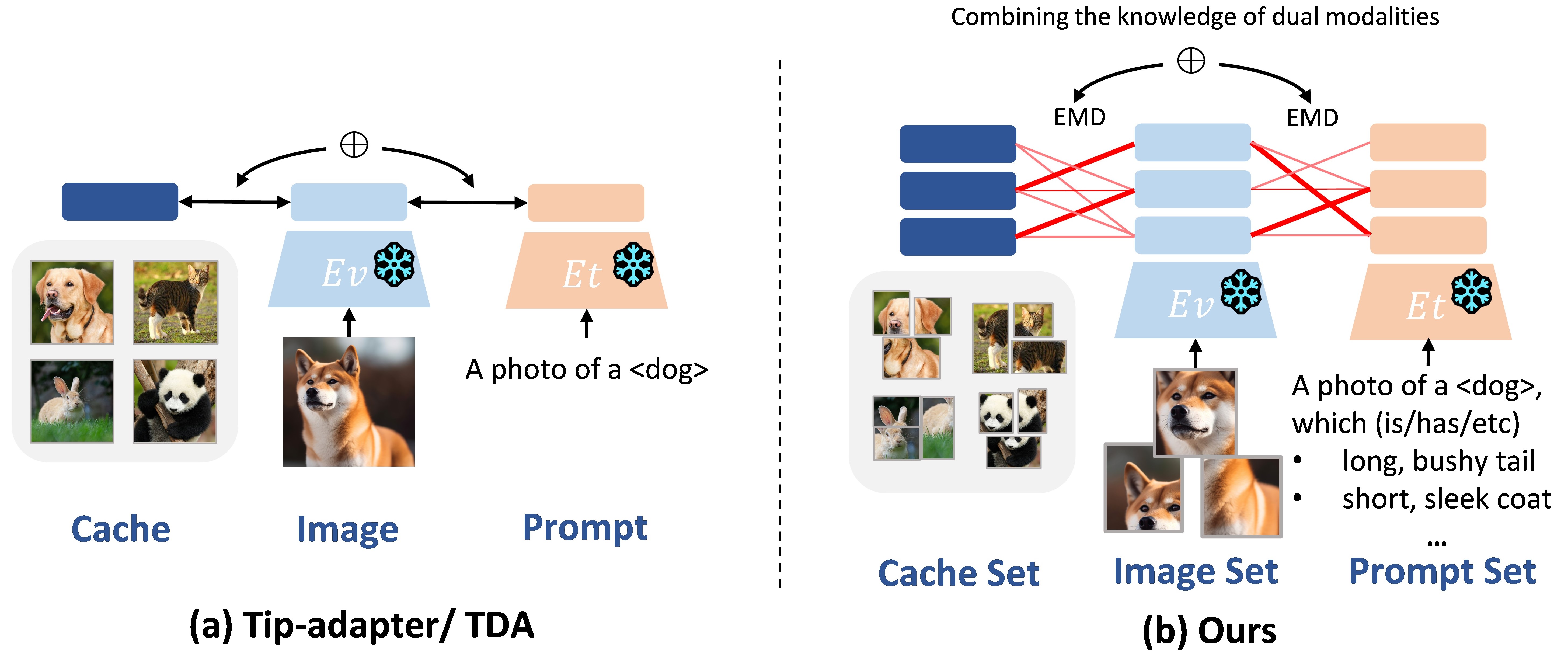} 
\caption{Overview of our proposed method under few-shot and test-time adaption scenarios. (a) Tip-Adapter and TDA constructs a non-parametric cache during training stage or testing stage. (b) Our method builds upon Tip-Adapter and TDA. We construct a local-aware cache and during classification, we use the Earth Mover's Distance (EMD) instead of cosine similarity.}
\label{fig:method2}
\end{figure*}
\textcolor{black}{\textbf{Tip-Adapter and TDA Preliminaries.}
Building on the CLIP-based adaptation paradigm, recent works such as Tip-Adapter \citep{zhang2021tip} and TDA \citep{karmanov2024efficient} have demonstrated remarkable efficiency in few-shot learning and test-time adaptation scenarios. As shown in Fig.~\ref{fig:method2}(a), these approaches construct a non-parametric cache during training stage or testing stage. For example, the Tip-Adapter builds a key-value cache model to store visual features of few-shot training images and their corresponding labels as key-value pairs. During inference, the features of test images are used as queries to retrieve relevant knowledge from the cache model. While effective, these methods primarily operate on global image representations, potentially overlooking fine-grained local patterns that could enhance discriminative power.}

\textcolor{black}{\textbf{Constructing local-aware caches.}
The proposed plug-and-play solution enhances CLIP's local perception abilities, allowing us to reformulate the cache construction for Tip-Adapter to further improve its performance. The similar approach can also be applied to TDA. Given a training set \(\mathcal{D}_{\text{train}} = \{(x_i, y_i)\}_{i=1}^S\) with \(S\) labeled examples, we perform \(H\) random crops for each image \(x_i\). Let $I_c={i_1,i_2,...,i_K}$ denote the descriptor-only prompt set for class $c$. For each descriptor \(i_k\), we select the visual feature \(v_{i,h}\) from the cropped regions that has the highest cosine similarity to \(i_k\). These selected visual features are then used to construct a new cache for each class, providing a more precise representation for comparison with test images. This approach significantly improves the performance of Tip-Adapter and TDA by ensuring that the cache reflects the most important visual characteristics.}

After selecting the nearest visual feature for each text feature, we aggregate these \( K \) selected visual features into a set \( \mathbf{V}_i \) for each training image \( x_i \). For each class \( c \in \{1, \dots, C\} \), we construct a class-specific cache \( \mathcal{M}_c \) by collecting these tensors from all training images belonging to that class. 

\textbf{Integrating Knowledge from Dual Modalities}. As shown in Fig.~\ref{fig:method2}(b), building upon this cache mechanism, we refine the adapter's weight computation by replacing cosine similarity with EMD to better align local details. 
Let a test image $x$ be cropped into $M$ local regions, denoted as the set \( X = \{v_1, v_2, \dots, v_M\} \). For each class \( c \), we compute the Earth Mover's Distance (EMD) between the local feature set  \( X \) and the feature sets stored in the class-specific cache \( \mathcal{M}_c \). Specifically, we calculate the sum of EMD distances between \( X \) and each feature set in \( \mathcal{M}_c \):
\begin{equation}
\text{EMD}_c = \sum_{\mathbf{V} \in \mathcal{M}_c} \text{EMD}(X, \mathbf{V})
\end{equation}
The affinity score for each class \( c \) is then computed as:
\begin{equation}
A_c = \exp\left(-\beta \cdot \text{EMD}_c\right),
\end{equation}
where \( \beta > 0 \) is a hyperparameter that controls the sharpness of the affinity distribution.
To leverage both visual and textual information, we combine the affinity score \( A_c \) with the zero-shot textual similarity score. Then the final prediction can be made by selecting the class with highest fused similarity score: 
\begin{equation}
    \argmax_{c \in C}  \alpha \cdot A_c-\text{EMD}(X, D_c) .
\end{equation}

\section{Experiments}
\begin{table*}[ht]
    \centering
    \caption{Results on Zero-shot Classification Benchmark with ResNet-50 backbone. We report top-1 accuracy and “Average" is calculated by taking the mean accuracy across all 11 datasets. }
    \resizebox{2.05\columnwidth}{!}{
        \begin{tabular}{lccccccccccccc}
        \toprule
        \textbf{Method}  & \textbf{Pets} & \textbf{Flowers} & \textbf{FGVC} & \textbf{DTD} & \textbf{EuroSAT} & \textbf{Cars} & \textbf{Food} & \textbf{SUN} & \textbf{Caltech} & \textbf{UCF} & \textbf{ImageNet} & \textbf{Average} \\
        \midrule
        \textbf{Zero-Shot CLIP \cite{radford2021learning}}  & 85.77 & 66.14 & 17.28 & 42.32 & 37.56 & 55.61 & 77.31 & 58.52 & 86.29 & 61.46 & 58.18 & 58.77 \\
        \textbf{CALIP \cite{guo2023calip}} & 86.21 & 66.38 & 17.76 & 42.39 & \textbf{38.90} & 56.27 & 77.42 & 58.59 & 87.71 & 61.72 & 60.57 & 59.45 \\
        \textbf{CLIP+D\&D}  & \textbf{87.21} & \textbf{68.01} & \textbf{19.11} & \textbf{43.30} & 36.58 & \textbf{59.37} & \textbf{77.79} & \textbf{63.52} & \textbf{88.25} & \textbf{62.32} & \textbf{61.31} & \textbf{60.62} \\

        \bottomrule
        \end{tabular}
    }
    
    \label{tab:zero_shot}
\end{table*}
\begin{table*}[ht]
    \centering
    \caption{Results on Few-shot Learning Benchmark with ResNet-50 backbone. We report top-1 accuracy and "Average" is calculated by taking the mean accuracy across all 11 datasets. We report the 16-shot results under few-shot learning scenario.}
    \resizebox{2.05\columnwidth}{!}{
        \begin{tabular}{lcccccccccccc}
        \toprule
        \textbf{Method}  & \textbf{Pets} & \textbf{Flowers} & \textbf{FGVC} & \textbf{DTD} & \textbf{EuroSAT} & \textbf{Cars} & \textbf{Food} & \textbf{SUN} & \textbf{Caltech} & \textbf{UCF} & \textbf{ImageNet} & \textbf{Average} \\
        \midrule
        \textbf{Tip-Adapter \cite{zhang2021tip}}  & 88.14 & 89.89 & 29.76 & 60.93 & 70.54 & 66.77 & 77.83 & 66.85 & 90.18 & 70.58 & 62.01 & 70.32 \\
        \textbf{Linear-probe \cite{radford2021learning}}  & 76.42 & \textbf{94.95} & \textbf{36.39} & \textbf{63.97} & \textbf{82.76} & \textbf{70.08} & 70.17 & 67.15 & 90.63 & \textbf{73.72} & 55.87 & 71.10 \\
        \textbf{Tip-X \cite{udandarao2023sus}}  & \textbf{89.86} & 90.29 & 30.12 & 63.53 & 73.12 & 67.30 & 77.93 & 68.00 & 90.70 & 71.95 & 62.61 & 71.40 \\
        \textbf{Tip+D\&D} & 88.36 & 90.17 & 30.50 & 62.16 & 72.47 & 68.91 & \textbf{78.46} & \textbf{68.81} & \textbf{91.13} & 72.40 & \textbf{62.74} & \textbf{71.46} \\
        \bottomrule
        \end{tabular}
    }
    \label{tab:few_shot}
\end{table*}
\begin{table*}[htbp]
  \centering
  \caption{Full results on the Cross-Domain Benchmark with ViT-B/16 backbone. We report top-1 accuracy and “Average" is calculated by taking the mean accuracy across all ten datasets.}
  \resizebox{2.05\columnwidth}{!}{
    \begin{tabular}{l|cccccccccc|c}
    \toprule
          & \textbf{Caltech} & \textbf{Pets} & \textbf{Cars} & \textbf{Flowers} & \textbf{Food101} & \textbf{Aircraft} & \textbf{SUN397} & \textbf{DTD} & \textbf{EuroSAT} & \textbf{UCF101} & \textit{Average} \\
    \midrule
    \textbf{CLIP \cite{radford2021learning}} & 93.35  & 88.25 & 65.48 & 67.44 & 83.65 & 23.67 & 62.59 & 44.27 & 42.01 & 65.13 & 63.58  \\
    \textbf{CLIP + TPT \cite{shu2022test}} & 94.16  & 87.79  & 66.87  & 68.98  & 84.67  & 24.78  & 65.50  & 47.75  & 42.44  & 68.04  & 65.10  \\
    \textbf{MaPLe+TPT \cite{shu2022test}} & 93.59  & 90.72  & 66.50  & 72.37  & 86.64  & 24.70  & 67.54  & 45.87  & 47.80  & 69.19  & 66.50  \\
    \textbf{DiffTPT \cite{feng2023diverse}} & 92.49  & 88.22  & 67.01  & 70.10  & 87.23 & 25.60  & 65.74  & 47.00  & 43.13  & 62.67  & 65.47  \\
    \textbf{PromptAlign \cite{abdul2023align}} & 94.01  &\textbf{90.76} & 68.50  & 72.39 & 86.65  & 24.80  & 67.54  & 47.24  & 47.86  & 69.47  & 66.92  \\
    \textbf{TDA \cite{karmanov2024efficient}} & 94.24  & 88.63  & 67.28  & 71.42  & 86.14  & 23.91  & 67.62  & 47.40  & 58.00  & 70.66  & 67.53  \\
    \midrule
    \textbf{TDA+D\&D} & \textbf{94.97}  & 90.68 & \textbf{68.72}  & \textbf{72.76}  & \textbf{87.26}  & \textbf{27.87} & \textbf{67.98}  & \textbf{52.19} & \textbf{59.21}  & \textbf{73.12} & \textbf{69.48} \\
    \bottomrule
    \end{tabular}%
    }
  \label{tab:tta}%
\end{table*}%
\subsection{Implementation details}
\textbf{Dataset setup.} \textcolor{black}{In this study, we use 11 commonly used image classification datasets, including ImageNet \cite{deng2009imagenet}, Caltech101 \cite{fei2004learning}, OxfordPets \cite{parkhi2012cats}, StanfordCars \cite{krause20133d}, Flowers102 \cite{nilsback2008automated}, Food101 \cite{bossard2014food}, EuroSAT \cite{helber2019eurosat}, UCF101 \cite{soomro2012ucf101}, SUN397 \cite{xiao2010sun}, FGVC \cite{maji2013fine}and DTD \cite{cimpoi2014describing}. These datasets cover a range of visual recognition tasks, providing a solid benchmark for evaluating model performance across different domains.}\\
\textbf{Experimental setup.} We conducted experiments under three distinct scenarios to evaluate the model’s generalization and adaptability: zero-shot CLIP, few-shot learning scenarios, and test-time adaptation scenarios. The experimental configurations for each scenario are detailed as follows:
\textbf{Zero-shot Classification.} The pre-trained CLIP model (ResNet-50 backbone) was utilized without any fine-tuning. To ensure a reliable estimate of performance, we conduct three independent runs with different random seeds and report the average results.
\textbf{Few-shot Learning.} Following the evaluation protocol proposed by CLIP \cite{radford2021learning}, for each \(M\)-shot scenario, we randomly select \(M\) instances per class, where \(M \in \{1, 2, 4, 8, 16\}\). We then train the model on these selected samples and evaluate its performance on the full test set. In our experiments, we conduct three independent runs with different random seeds and report the average results. We adopt the same hyperparameter search settings as Tip-Adapter~\cite{zhang2021tip}.
\textbf{Test-time Adaptation.} In test-time adaptation, the batch size is set to be 1. We adopt the same hyperparameters as TDA~\cite{karmanov2024efficient}.

All our experiments are conducted with a Nvidia 4090 24GB GPU. For computational convenience, each input image was cropped into $M = \text{9}$ patches through random cropping. Similarly, we provided 9 descriptions for each class. To efficiently approximate the Earth Mover's Distance (EMD), we employed the Sinkhorn distance, which offers a fast and accurate estimation of EMD.
\subsection{Experimental Results}
\textbf{Zero-shot Classification.}

As shown in Tab.~\ref{tab:zero_shot}, we compare zero-shot performance with CLIP \cite{radford2021learning} and CALIP \cite{guo2023calip} models for all 11 datasets. Our method achieves superior zero-shot performance across 11 datasets , outperforming Zero-Shot CLIP and CALIP by 1.85\% and 1.17\% respectively. Key improvements include +1.71\% on fine-grained tasks (Pets/Flowers/FGVC) and +5\% on SUN-397, demonstrating enhanced cross-modal alignment. Our method maintains stronger generalization on mainstream benchmarks like ImageNet (59.68\% vs 61.31\%). Results validate our approach’s effectiveness in extracting local features for set-based classification.\\
\textbf{Few-shot Learning.}

As shown in the Tab.~\ref{tab:few_shot}, we further analyze the performance of our method on the Few-Shot Learning benchmark.  Compared to existing adapter-based approaches, Tip+D\&D achieves the highest average performance of 71.46\%.  Our method demonstrates consistent improvements on all 11 datasets, compared to the original Tip-adapter.  While Linear-probe excels in some datasets, Tip+D\&D outperforms it in other key areas, proving to be a more well-rounded method. In summary, Tip+D\&D shows superior consistency and higher overall accuracy across most evaluation metrics. The complete results for 1, 2, 4, 8, and 16 shots can be found in Appendix \ref{appendix:few_all}.\\
\textbf{Test-time Adaptation.}

We further highlight the enhanced transfer capability of our method on the Cross-Domain benchmark and present the results in Tab.~\ref{tab:tta}. Compared to both training-based and training-free approaches, TDA+D\&D achieves state-of-the-art average performance, surpassing the strongest baseline (TDA) by 1.95\% and outperforming existing methods on 9 out of 10 tasks. By adaptively integrating multi-scale feature representations, our method effectively captures discriminative knowledge to distinguish fine-grained classes. Notably, for datasets requiring detailed differentiation, such as Aircraft, our approach attains a significant improvement of 27.87\%, which outperforms all baselines. Additionally, we observe remarkable gains on DTD and UCF101 , demonstrating robust generalization across diverse domains. The consistent improvements underscore the efficacy of our adaptive framework in addressing both coarse and fine-grained recognition challenges. More results with ResNet-50 backbone can be found in Appendix \ref{appendix:robustness}. 
\subsection{Ablation Study}
In this section, we primarily focus on issues that may be contentious, in order to illustrate the effectiveness of our approach.\\
\textbf{Q1. Core contribution isolation.}  How can we systematically isolate and assess whether the performance improvements of our method are due to our core contribution, rather than the added textual descriptions?
\begin{table}[htbp]
  \centering
  \caption{Ablation studies under zero-shot scenario. \textbf{D}: Enhancing CLIP with detailed textual descriptions, \textbf{R}: Generating image sets through random cropping and averaging their features.}
    \begin{tabular}{c|ccc}
    \toprule
    \textbf{Method} & \textbf{FGVC} & \textbf{UCF} & \textbf{ImageNet} \\
    \midrule
    \textbf{CLIP} & 17.28 & 61.46 & 58.18 \\
    \textbf{CLIP+D} & 17.16 & 61.99 & 60.02 \\
    \textbf{CLIP+D+R} & 18.86 & 61.99 & 61.17 \\
    \textbf{CLIP+D\&D} & \textbf{19.11} & \textbf{62.32} & \textbf{61.31} \\
    \bottomrule
    \end{tabular}%
  \label{tab:abla}%
\end{table}%

To isolate our core contribution from textual augmentation, we conducted an ablation study on three benchmarks: UCF101 for action recognition (minimal text advantage), FGVC for fine-grained classification (subtle attribute localization), and ImageNet as a large-scale reference (isolated text correlations).

As shown in Tab.~\ref{tab:abla}, results show that textual description augmentation (D) significantly improves CLIP’s performance on generic classification tasks like ImageNet (+1.84\%), yet yields marginal gains on fine-grained recognition (FGVC, -0.12\%). This is because conventional text augmentation merely appends generic details to class names (e.g., "bird" → "bird with white-spotted wings"), which strengthens overall class semantics but fails to discriminate subtle inter-class distinctions (e.g., between bird subspecies). Our method addresses this by applying random cropping to images and computing Earth Mover’s Distance (EMD) with prompt set. This forces the model to align fine-grained local features with diverse textual cues, achieving a +1.95\% improvement over CLIP+D on FGVC. 
\\
\textbf{Q2. Alternative classification methods.} Are there other methods for classification regarding the randomly cropped image set and prompt set? (e.g., calculating the average for both the image set and prompt set, and then classifying based on cosine similarity)

As shown in Tab.~\ref{tab:abla}, traditional random cropping with feature averaging (CLIP+D+R) shows no gains on UCF, as averaging dilutes spatiotemporal context critical for action recognition (e.g., "dribbling a ball" requires correlating hand and leg motions). In contrast, our method leverages EMD to compute optimal alignment between image crops and text descriptions, enabling cross-modal local matching. The EMD-based weighted matching preserves action coherence, leading to superior performance on UCF (62.32\% vs. 61.99\%). Overall, our method consistently outperforms CLIP+D+R across all datasets, demonstrating significant advantages in enhancing fine-grained classification and retaining local information.

\section{Conclusion and Future Work}
Our work reveals CLIP’s inherent bias toward global semantics and its underutilized capacity for local descriptor understanding. To address this issue, we introduce a plug-and-play adaptation method that reactivates this potential through random cropping, thereby enabling CLIP to capture both global and local features. Extensive experiments under zero-shot, few-shot, and test-time adaptation settings achieves state-of-the-art results compared to previous methods. These results validate the effectiveness of our method. However, there is still room for improvement. Future work may explore integrating semantic segmentation models like SAM to enable precise localization of descriptor-relevant areas (e.g., extracting beak regions guided by text), further enhancing the alignment between visual and textual information. Additionally, considering that CLIP’s pretraining objective currently lacks explicit supervision for part-text alignment, future frameworks could co-train VLMs with localized contrastive losses.
{
    \small
    \bibliographystyle{ieeenat_fullname}
    \bibliography{main}

\begin{thebibliography}{50}
\providecommand{\natexlab}[1]{#1}
\providecommand{\url}[1]{\texttt{#1}}
\expandafter\ifx\csname urlstyle\endcsname\relax
  \providecommand{\doi}[1]{doi: #1}\else
  \providecommand{\doi}{doi: \begingroup \urlstyle{rm}\Url}\fi

\bibitem[Abdul~Samadh et~al.(2023)Abdul~Samadh, Gani, Hussein, Khattak, Naseer, Shahbaz~Khan, and Khan]{abdul2023align}
Jameel Abdul~Samadh, Mohammad~Hanan Gani, Noor Hussein, Muhammad~Uzair Khattak, Muhammad~Muzammal Naseer, Fahad Shahbaz~Khan, and Salman~H Khan.
\newblock Align your prompts: Test-time prompting with distribution alignment for zero-shot generalization.
\newblock \emph{Advances in Neural Information Processing Systems}, 36:\penalty0 80396--80413, 2023.

\bibitem[Arjovsky et~al.(2017)Arjovsky, Chintala, and Bottou]{arjovsky2017wasserstein}
Martin Arjovsky, Soumith Chintala, and L{\'e}on Bottou.
\newblock Wasserstein generative adversarial networks.
\newblock In \emph{International conference on machine learning}, pages 214--223. PMLR, 2017.

\bibitem[Bossard et~al.(2014)Bossard, Guillaumin, and Van~Gool]{bossard2014food}
Lukas Bossard, Matthieu Guillaumin, and Luc Van~Gool.
\newblock Food-101--mining discriminative components with random forests.
\newblock In \emph{Computer vision--ECCV 2014: 13th European conference, zurich, Switzerland, September 6-12, 2014, proceedings, part VI 13}, pages 446--461. Springer, 2014.

\bibitem[Brown et~al.(2020)Brown, Mann, Ryder, Subbiah, Kaplan, Dhariwal, Neelakantan, Shyam, Sastry, Askell, et~al.]{brown2020language}
Tom Brown, Benjamin Mann, Nick Ryder, Melanie Subbiah, Jared~D Kaplan, Prafulla Dhariwal, Arvind Neelakantan, Pranav Shyam, Girish Sastry, Amanda Askell, et~al.
\newblock Language models are few-shot learners.
\newblock \emph{Advances in neural information processing systems}, 33:\penalty0 1877--1901, 2020.

\bibitem[Bunne et~al.(2024)Bunne, Schiebinger, Krause, Regev, and Cuturi]{bunne2024optimal}
Charlotte Bunne, Geoffrey Schiebinger, Andreas Krause, Aviv Regev, and Marco Cuturi.
\newblock Optimal transport for single-cell and spatial omics.
\newblock \emph{Nature Reviews Methods Primers}, 4\penalty0 (1):\penalty0 58, 2024.

\bibitem[Choi et~al.(2024)Choi, Choi, and Kang]{choi2024scalable}
Jaemoo Choi, Jaewoong Choi, and Myungjoo Kang.
\newblock Scalable wasserstein gradient flow for generative modeling through unbalanced optimal transport.
\newblock In \emph{Proceedings of the 41st International Conference on Machine Learning}, pages 8629--8650, 2024.

\bibitem[Cimpoi et~al.(2014)Cimpoi, Maji, Kokkinos, Mohamed, and Vedaldi]{cimpoi2014describing}
Mircea Cimpoi, Subhransu Maji, Iasonas Kokkinos, Sammy Mohamed, and Andrea Vedaldi.
\newblock Describing textures in the wild.
\newblock In \emph{Proceedings of the IEEE conference on computer vision and pattern recognition}, pages 3606--3613, 2014.

\bibitem[Courty et~al.(2016)Courty, Flamary, Tuia, and Rakotomamonjy]{courty2016optimal}
Nicolas Courty, R{\'e}mi Flamary, Devis Tuia, and Alain Rakotomamonjy.
\newblock Optimal transport for domain adaptation.
\newblock \emph{IEEE transactions on pattern analysis and machine intelligence}, 39\penalty0 (9):\penalty0 1853--1865, 2016.

\bibitem[Deng et~al.(2009)Deng, Dong, Socher, Li, Li, and Fei-Fei]{deng2009imagenet}
Jia Deng, Wei Dong, Richard Socher, Li-Jia Li, Kai Li, and Li Fei-Fei.
\newblock Imagenet: A large-scale hierarchical image database.
\newblock In \emph{2009 IEEE conference on computer vision and pattern recognition}, pages 248--255. Ieee, 2009.

\bibitem[Ding et~al.(2022)Ding, Xue, Xia, and Dai]{ding2022decoupling}
Jian Ding, Nan Xue, Gui-Song Xia, and Dengxin Dai.
\newblock Decoupling zero-shot semantic segmentation.
\newblock In \emph{Proceedings of the IEEE/CVF conference on computer vision and pattern recognition}, pages 11583--11592, 2022.

\bibitem[Fatras et~al.(2021)Fatras, S{\'e}journ{\'e}, Flamary, and Courty]{fatras2021unbalanced}
Kilian Fatras, Thibault S{\'e}journ{\'e}, R{\'e}mi Flamary, and Nicolas Courty.
\newblock Unbalanced minibatch optimal transport; applications to domain adaptation.
\newblock In \emph{International Conference on Machine Learning}, pages 3186--3197. PMLR, 2021.

\bibitem[Fei-Fei et~al.(2004)Fei-Fei, Fergus, and Perona]{fei2004learning}
Li Fei-Fei, Rob Fergus, and Pietro Perona.
\newblock Learning generative visual models from few training examples: An incremental bayesian approach tested on 101 object categories.
\newblock In \emph{2004 conference on computer vision and pattern recognition workshop}, pages 178--178. IEEE, 2004.

\bibitem[Feng et~al.(2023)Feng, Yu, Liu, Khan, and Zuo]{feng2023diverse}
Chun-Mei Feng, Kai Yu, Yong Liu, Salman Khan, and Wangmeng Zuo.
\newblock Diverse data augmentation with diffusions for effective test-time prompt tuning.
\newblock In \emph{Proceedings of the IEEE/CVF International Conference on Computer Vision}, pages 2704--2714, 2023.

\bibitem[Gao et~al.(2024)Gao, Geng, Zhang, Ma, Fang, Zhang, Li, and Qiao]{gao2024clip}
Peng Gao, Shijie Geng, Renrui Zhang, Teli Ma, Rongyao Fang, Yongfeng Zhang, Hongsheng Li, and Yu Qiao.
\newblock Clip-adapter: Better vision-language models with feature adapters.
\newblock \emph{International Journal of Computer Vision}, 132\penalty0 (2):\penalty0 581--595, 2024.

\bibitem[Gu et~al.()Gu, Lin, Kuo, and Cui]{guopen}
Xiuye Gu, Tsung-Yi Lin, Weicheng Kuo, and Yin Cui.
\newblock Open-vocabulary object detection via vision and language knowledge distillation.
\newblock In \emph{International Conference on Learning Representations}.

\bibitem[Guo et~al.(2023)Guo, Zhang, Qiu, Ma, Miao, He, and Cui]{guo2023calip}
Ziyu Guo, Renrui Zhang, Longtian Qiu, Xianzheng Ma, Xupeng Miao, Xuming He, and Bin Cui.
\newblock Calip: Zero-shot enhancement of clip with parameter-free attention.
\newblock In \emph{Proceedings of the AAAI Conference on Artificial Intelligence}, pages 746--754, 2023.

\bibitem[Gupta et~al.(2022)Gupta, Narayan, Joseph, Khan, Khan, and Shah]{gupta2022ow}
Akshita Gupta, Sanath Narayan, KJ Joseph, Salman Khan, Fahad~Shahbaz Khan, and Mubarak Shah.
\newblock Ow-detr: Open-world detection transformer.
\newblock In \emph{Proceedings of the IEEE/CVF conference on computer vision and pattern recognition}, pages 9235--9244, 2022.

\bibitem[Helber et~al.(2019)Helber, Bischke, Dengel, and Borth]{helber2019eurosat}
Patrick Helber, Benjamin Bischke, Andreas Dengel, and Damian Borth.
\newblock Eurosat: A novel dataset and deep learning benchmark for land use and land cover classification.
\newblock \emph{IEEE Journal of Selected Topics in Applied Earth Observations and Remote Sensing}, 12\penalty0 (7):\penalty0 2217--2226, 2019.

\bibitem[Jia et~al.(2021)Jia, Yang, Xia, Chen, Parekh, Pham, Le, Sung, Li, and Duerig]{jia2021scaling}
Chao Jia, Yinfei Yang, Ye Xia, Yi-Ting Chen, Zarana Parekh, Hieu Pham, Quoc Le, Yun-Hsuan Sung, Zhen Li, and Tom Duerig.
\newblock Scaling up visual and vision-language representation learning with noisy text supervision.
\newblock In \emph{International conference on machine learning}, pages 4904--4916. PMLR, 2021.

\bibitem[Joseph et~al.(2021)Joseph, Khan, Khan, and Balasubramanian]{joseph2021towards}
KJ Joseph, Salman Khan, Fahad~Shahbaz Khan, and Vineeth~N Balasubramanian.
\newblock Towards open world object detection.
\newblock In \emph{Proceedings of the IEEE/CVF conference on computer vision and pattern recognition}, pages 5830--5840, 2021.

\bibitem[Karmanov et~al.(2024)Karmanov, Guan, Lu, El~Saddik, and Xing]{karmanov2024efficient}
Adilbek Karmanov, Dayan Guan, Shijian Lu, Abdulmotaleb El~Saddik, and Eric Xing.
\newblock Efficient test-time adaptation of vision-language models.
\newblock In \emph{Proceedings of the IEEE/CVF Conference on Computer Vision and Pattern Recognition}, pages 14162--14171, 2024.

\bibitem[Krause et~al.(2013)Krause, Stark, Deng, and Fei-Fei]{krause20133d}
Jonathan Krause, Michael Stark, Jia Deng, and Li Fei-Fei.
\newblock 3d object representations for fine-grained categorization.
\newblock In \emph{Proceedings of the IEEE international conference on computer vision workshops}, pages 554--561, 2013.

\bibitem[Li et~al.()Li, Weinberger, Belongie, Koltun, and Ranftl]{lilanguage}
Boyi Li, Kilian~Q Weinberger, Serge Belongie, Vladlen Koltun, and Rene Ranftl.
\newblock Language-driven semantic segmentation.
\newblock In \emph{International Conference on Learning Representations}.

\bibitem[Maji et~al.(2013)Maji, Rahtu, Kannala, Blaschko, and Vedaldi]{maji2013fine}
Subhransu Maji, Esa Rahtu, Juho Kannala, Matthew Blaschko, and Andrea Vedaldi.
\newblock Fine-grained visual classification of aircraft.
\newblock \emph{arXiv preprint arXiv:1306.5151}, 2013.

\bibitem[Mao et~al.(2023)Mao, Teotia, Sundar, Menon, Yang, Wang, and Vondrick]{mao2023doubly}
Chengzhi Mao, Revant Teotia, Amrutha Sundar, Sachit Menon, Junfeng Yang, Xin Wang, and Carl Vondrick.
\newblock Doubly right object recognition: A why prompt for visual rationales.
\newblock In \emph{Proceedings of the IEEE/CVF Conference on Computer Vision and Pattern Recognition}, pages 2722--2732, 2023.

\bibitem[Menon and Vondrick()]{menonvisual}
Sachit Menon and Carl Vondrick.
\newblock Visual classification via description from large language models.
\newblock In \emph{The Eleventh International Conference on Learning Representations}.

\bibitem[Monge(1781)]{monge1781memoire}
Gaspard Monge.
\newblock M{\'e}moire sur la th{\'e}orie des d{\'e}blais et des remblais.
\newblock \emph{Mem. Math. Phys. Acad. Royale Sci.}, pages 666--704, 1781.

\bibitem[Nilsback and Zisserman(2008)]{nilsback2008automated}
Maria-Elena Nilsback and Andrew Zisserman.
\newblock Automated flower classification over a large number of classes.
\newblock In \emph{2008 Sixth Indian conference on computer vision, graphics \& image processing}, pages 722--729. IEEE, 2008.

\bibitem[Parkhi et~al.(2012)Parkhi, Vedaldi, Zisserman, and Jawahar]{parkhi2012cats}
Omkar~M Parkhi, Andrea Vedaldi, Andrew Zisserman, and CV Jawahar.
\newblock Cats and dogs.
\newblock In \emph{2012 IEEE conference on computer vision and pattern recognition}, pages 3498--3505. IEEE, 2012.

\bibitem[Peyr{\'e} et~al.(2019)Peyr{\'e}, Cuturi, et~al.]{peyre2019computational}
Gabriel Peyr{\'e}, Marco Cuturi, et~al.
\newblock Computational optimal transport: With applications to data science.
\newblock \emph{Foundations and Trends{\textregistered} in Machine Learning}, 11\penalty0 (5-6):\penalty0 355--607, 2019.

\bibitem[Pratt et~al.(2023)Pratt, Covert, Liu, and Farhadi]{pratt2023does}
Sarah Pratt, Ian Covert, Rosanne Liu, and Ali Farhadi.
\newblock What does a platypus look like? generating customized prompts for zero-shot image classification.
\newblock In \emph{Proceedings of the IEEE/CVF International Conference on Computer Vision}, pages 15691--15701, 2023.

\bibitem[Radford et~al.(2021)Radford, Kim, Hallacy, Ramesh, Goh, Agarwal, Sastry, Askell, Mishkin, Clark, et~al.]{radford2021learning}
Alec Radford, Jong~Wook Kim, Chris Hallacy, Aditya Ramesh, Gabriel Goh, Sandhini Agarwal, Girish Sastry, Amanda Askell, Pamela Mishkin, Jack Clark, et~al.
\newblock Learning transferable visual models from natural language supervision.
\newblock In \emph{International conference on machine learning}, pages 8748--8763. PmLR, 2021.

\bibitem[Roth et~al.(2023)Roth, Kim, Koepke, Vinyals, Schmid, and Akata]{roth2023waffling}
Karsten Roth, Jae~Myung Kim, A Koepke, Oriol Vinyals, Cordelia Schmid, and Zeynep Akata.
\newblock Waffling around for performance: Visual classification with random words and broad concepts.
\newblock In \emph{Proceedings of the IEEE/CVF International Conference on Computer Vision}, pages 15746--15757, 2023.

\bibitem[Rubner et~al.(2000)Rubner, Tomasi, and Guibas]{rubner2000earth}
Yossi Rubner, Carlo Tomasi, and Leonidas~J Guibas.
\newblock The earth mover's distance as a metric for image retrieval.
\newblock \emph{International journal of computer vision}, 40:\penalty0 99--121, 2000.

\bibitem[Shen et~al.(2021)Shen, Feydy, Liu, Curiale, San Jose~Estepar, San Jose~Estepar, and Niethammer]{shen2021accurate}
Zhengyang Shen, Jean Feydy, Peirong Liu, Ariel~H Curiale, Ruben San Jose~Estepar, Raul San Jose~Estepar, and Marc Niethammer.
\newblock Accurate point cloud registration with robust optimal transport.
\newblock \emph{Advances in Neural Information Processing Systems}, 34:\penalty0 5373--5389, 2021.

\bibitem[Shu et~al.(2022)Shu, Nie, Huang, Yu, Goldstein, Anandkumar, and Xiao]{shu2022test}
Manli Shu, Weili Nie, De-An Huang, Zhiding Yu, Tom Goldstein, Anima Anandkumar, and Chaowei Xiao.
\newblock Test-time prompt tuning for zero-shot generalization in vision-language models.
\newblock \emph{Advances in Neural Information Processing Systems}, 35:\penalty0 14274--14289, 2022.

\bibitem[Soomro(2012)]{soomro2012ucf101}
K Soomro.
\newblock Ucf101: A dataset of 101 human actions classes from videos in the wild.
\newblock \emph{arXiv preprint arXiv:1212.0402}, 2012.

\bibitem[Thual et~al.(2022)Thual, Tran, Zemskova, Courty, Flamary, Dehaene, and Thirion]{thual2022aligning}
Alexis Thual, Quang~Huy Tran, Tatiana Zemskova, Nicolas Courty, R{\'e}mi Flamary, Stanislas Dehaene, and Bertrand Thirion.
\newblock Aligning individual brains with fused unbalanced gromov wasserstein.
\newblock \emph{Advances in neural information processing systems}, 35:\penalty0 21792--21804, 2022.

\bibitem[Touvron et~al.(2023)Touvron, Lavril, Izacard, Martinet, Lachaux, Lacroix, Rozi{\`e}re, Goyal, Hambro, Azhar, et~al.]{touvron2023llama}
Hugo Touvron, Thibaut Lavril, Gautier Izacard, Xavier Martinet, Marie-Anne Lachaux, Timoth{\'e}e Lacroix, Baptiste Rozi{\`e}re, Naman Goyal, Eric Hambro, Faisal Azhar, et~al.
\newblock Llama: Open and efficient foundation language models.
\newblock \emph{arXiv preprint arXiv:2302.13971}, 2023.

\bibitem[Udandarao et~al.(2023)Udandarao, Gupta, and Albanie]{udandarao2023sus}
Vishaal Udandarao, Ankush Gupta, and Samuel Albanie.
\newblock Sus-x: Training-free name-only transfer of vision-language models.
\newblock In \emph{Proceedings of the IEEE/CVF International Conference on Computer Vision}, pages 2725--2736, 2023.

\bibitem[Xiao et~al.(2010)Xiao, Hays, Ehinger, Oliva, and Torralba]{xiao2010sun}
Jianxiong Xiao, James Hays, Krista~A Ehinger, Aude Oliva, and Antonio Torralba.
\newblock Sun database: Large-scale scene recognition from abbey to zoo.
\newblock In \emph{2010 IEEE computer society conference on computer vision and pattern recognition}, pages 3485--3492. IEEE, 2010.

\bibitem[Xu and Gould(2024)]{xu2024temporally}
Ming Xu and Stephen Gould.
\newblock Temporally consistent unbalanced optimal transport for unsupervised action segmentation.
\newblock In \emph{Proceedings of the IEEE/CVF Conference on Computer Vision and Pattern Recognition}, pages 14618--14627, 2024.

\bibitem[Xu et~al.(2022)Xu, Zhang, Wei, Lin, Cao, Hu, and Bai]{xu2022simple}
Mengde Xu, Zheng Zhang, Fangyun Wei, Yutong Lin, Yue Cao, Han Hu, and Xiang Bai.
\newblock A simple baseline for open-vocabulary semantic segmentation with pre-trained vision-language model.
\newblock In \emph{European Conference on Computer Vision}, pages 736--753. Springer, 2022.

\bibitem[Xu et~al.(2020)Xu, Liu, Wang, Chen, and Wang]{xu2020reliable}
Renjun Xu, Pelen Liu, Liyan Wang, Chao Chen, and Jindong Wang.
\newblock Reliable weighted optimal transport for unsupervised domain adaptation.
\newblock In \emph{Proceedings of the IEEE/CVF conference on computer vision and pattern recognition}, pages 4394--4403, 2020.

\bibitem[Yuksekgonul et~al.()Yuksekgonul, Bianchi, Kalluri, Jurafsky, and Zou]{yuksekgonuland}
Mert Yuksekgonul, Federico Bianchi, Pratyusha Kalluri, Dan Jurafsky, and James Zou.
\newblock When and why vision-language models behave like bags-of-words, and what to do about it?
\newblock In \emph{The Eleventh International Conference on Learning Representations}.

\bibitem[Zhang et~al.(2020)Zhang, Cai, Lin, and Shen]{zhang2020deepemd}
Chi Zhang, Yujun Cai, Guosheng Lin, and Chunhua Shen.
\newblock Deepemd: Few-shot image classification with differentiable earth mover's distance and structured classifiers.
\newblock In \emph{Proceedings of the IEEE/CVF conference on computer vision and pattern recognition}, pages 12203--12213, 2020.

\bibitem[Zhang et~al.(2021)Zhang, Fang, Zhang, Gao, Li, Dai, Qiao, and Li]{zhang2021tip}
Renrui Zhang, Rongyao Fang, Wei Zhang, Peng Gao, Kunchang Li, Jifeng Dai, Yu Qiao, and Hongsheng Li.
\newblock Tip-adapter: Training-free clip-adapter for better vision-language modeling.
\newblock \emph{arXiv preprint arXiv:2111.03930}, 2021.

\bibitem[Zhao et~al.()Zhao, Phung, Huynh, Le, and Buntine]{zhaoneural}
He Zhao, Dinh Phung, Viet Huynh, Trung Le, and Wray Buntine.
\newblock Neural topic model via optimal transport.
\newblock In \emph{International Conference on Learning Representations}.

\bibitem[Zhou et~al.(2022{\natexlab{a}})Zhou, Yang, Loy, and Liu]{zhou2022conditional}
Kaiyang Zhou, Jingkang Yang, Chen~Change Loy, and Ziwei Liu.
\newblock Conditional prompt learning for vision-language models.
\newblock In \emph{Proceedings of the IEEE/CVF conference on computer vision and pattern recognition}, pages 16816--16825, 2022{\natexlab{a}}.

\bibitem[Zhou et~al.(2022{\natexlab{b}})Zhou, Yang, Loy, and Liu]{zhou2022learning}
Kaiyang Zhou, Jingkang Yang, Chen~Change Loy, and Ziwei Liu.
\newblock Learning to prompt for vision-language models.
\newblock \emph{International Journal of Computer Vision}, 130\penalty0 (9):\penalty0 2337--2348, 2022{\natexlab{b}}.

\end{thebibliography}
}
\clearpage \appendix \section{Appendix}

\subsection{Building Descriptors}
\label{appendix:build}
\begin{figure*}[htbp]
\centering
\includegraphics[width=1\textwidth]{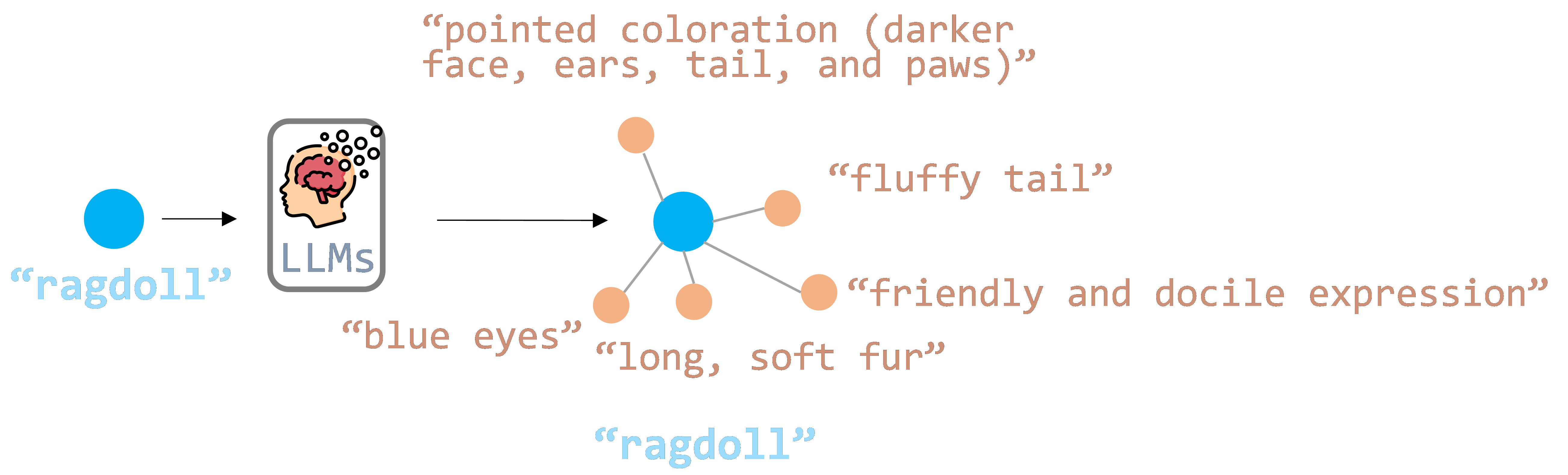} 
\caption{Detailed workflow to build descriptors.}
\label{fig:LLM}
\end{figure*}

We follow the method proposed in \cite{menonvisual} to build descriptors and the detailed workflow is illustrated in the Fig. \ref{fig:LLM}. Specifically, our prompt structure is of the form:\\
\begin{description}
\fontfamily{lmtt}\selectfont
\item[Q:] What are useful features for distinguishing a {category name} in a photo?
\item[A:] There are \{num\} useful visual features to tell there is a {category name} in a photo:
-
\end{description}

\begin{figure*}[htbp]
\centering
\includegraphics[width=1\textwidth]{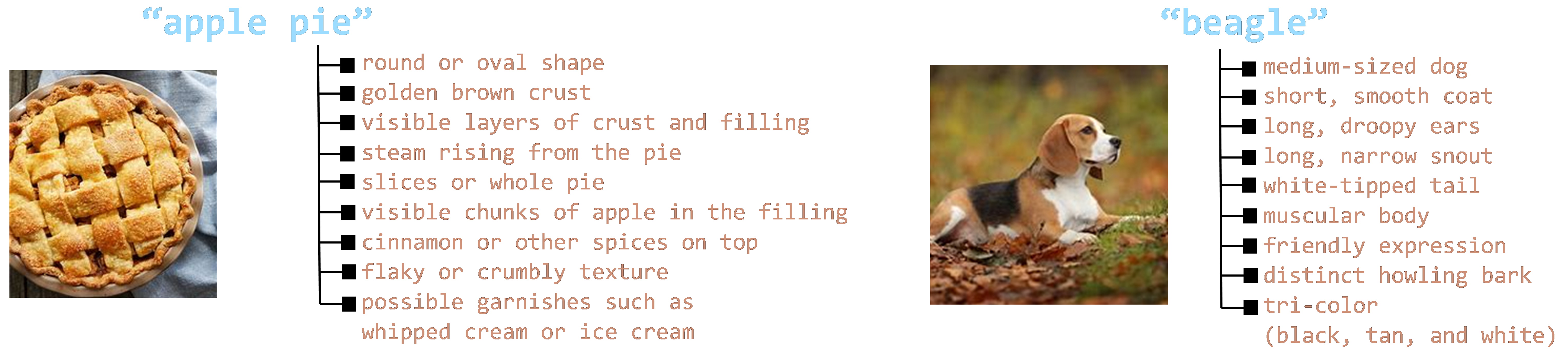} 
\caption{Examples of generated descriptors.}
\label{fig:des}
\end{figure*}

Where \{num\} refers to the number of descriptors you wish to obtain.\\
Here, we will demonstrate some examples of generated descriptors. The examples are shown in Fig. \ref{fig:des}

\subsection{Full Results on few-shot learning}
\label{appendix:few_all}
In this section, we will present the full results under few-shot learning scenario. As shown in Tab. \ref{tab:few_shot_all}, the results indicate that Tip+D\&D consistently outperforms Tip-adapter across all datasets and shot settings. This superior performance demonstrates the effectiveness of Tip+D\&D in adapting to new tasks with limited supervision, highlighting its potential for practical applications where data is scarce.

\subsection{Robustness to other Architectures}
\label{appendix:robustness}
To ensure the robustness of our proposed method, we evaluate the robustness of our method across different CLIP architectures under test-time adaption scenario. As shown in Tab.~\ref{tab:tta_rn50}, our method demonstrates superior performance, achieving the highest average accuracy of 62.25\%. Specifically, it outperforms other state-of-the-art methods in several datasets, such as EuroSAT (48.37\%) and UCF101 (65.61\%), and also shows strong results in SUN397 (63.05\%) and Flowers (68.09\%). This indicates that the proposed method effectively enhances robustness across diverse datasets, achieving a balanced improvement in overall performance compared to existing approaches.

\begin{table*}[htbp]
  \centering
  \caption{Full results under few-shot learning scenario.  Tip+D\&D consistently outperforms Tip across all datasets and shot settings.}
  \resizebox{2.1\columnwidth}{!}{
    \begin{tabular}{c|ccccccccccc|c}
    \toprule
    \textbf{Datasets} & \begin{sideways}\textbf{Caltech101}\end{sideways} & \begin{sideways}\textbf{Oxford Pets}\end{sideways} & \begin{sideways}\textbf{Stanford Cars}\end{sideways} & \begin{sideways}\textbf{Oxford Flowers}\end{sideways} & \begin{sideways}\textbf{Food101}\end{sideways} & \begin{sideways}\textbf{FGVC Aircraft}\end{sideways} & \begin{sideways}\textbf{SUN397}\end{sideways} & \begin{sideways}\textbf{DTD}\end{sideways} & \begin{sideways}\textbf{UCF101}\end{sideways} & \begin{sideways}\textbf{EuroSAT}\end{sideways} & \begin{sideways}\textbf{ImageNet}\end{sideways} & \textit{\textbf{Average}} \\
    \midrule
    \multicolumn{13}{c}{\textbf{1 shot}} \\
    \midrule
    \textbf{Tip} & 87.18  & 86.10  & 57.54  & 73.12  & 77.42  & 19.05  & 61.30  & 46.22  & 62.60  & 54.38  & 60.70  & 62.33  \\
    \textbf{Tip+D\&D} & \textbf{89.14} & \textbf{87.46} & \textbf{60.42} & \textbf{75.84} & \textbf{78.15} & \textbf{21.44} & \textbf{65.16} & \textbf{47.38} & \textbf{64.90} & \textbf{57.86} & \textbf{61.60} & \textbf{64.49} \\
    \midrule
    \multicolumn{13}{c}{\textbf{2 shots}} \\
    \midrule
    \textbf{Tip} & 88.44  & 87.03  & 57.93  & 79.13  & 77.52  & 21.21  & 62.70  & 49.47  & 64.74  & 61.68  & 60.96  & 64.62  \\
    \textbf{Tip+D\&D} & \textbf{89.52} & \textbf{87.97} & \textbf{62.20} & \textbf{80.36} & \textbf{78.37} & \textbf{22.21} & \textbf{66.19} & \textbf{50.85} & \textbf{66.16} & \textbf{65.24} & \textbf{61.77} & \textbf{66.44} \\
    \midrule
    \multicolumn{13}{c}{\textbf{4 shots}} \\
    \midrule
    \textbf{Tip} & 89.39  & 86.45  & 61.45  & 83.80  & 77.54  & 22.41  & 64.15  & 53.96  & 66.46  & 65.32  & 60.98  & 66.54  \\
    \textbf{Tip+D\&D} & \textbf{89.87} & \textbf{87.90} & \textbf{64.50} & \textbf{83.72} & \textbf{78.37} & \textbf{24.63} & \textbf{67.33} & \textbf{54.37} & \textbf{68.19} & \textbf{68.14} & \textbf{62.00} & \textbf{68.09} \\
    \midrule
    \multicolumn{13}{c}{\textbf{8 shots}} \\
    \midrule
    \textbf{Tip} & 89.83  & 87.03  & 62.93  & \textbf{87.98 } & 77.76  & 25.59  & 65.62  & 58.63  & 68.68  & 67.95  & 61.45  & 68.50  \\
    \textbf{Tip+D\&D} & \textbf{90.64} & \textbf{88.08} & \textbf{66.30} & 87.01  & \textbf{78.38} & \textbf{26.45} & \textbf{68.21} & \textbf{59.44} & \textbf{70.70} & \textbf{71.85} & \textbf{62.45} & \textbf{69.96} \\
    \midrule
    \multicolumn{13}{c}{\textbf{16 shots}} \\
    \midrule
    \textbf{Tip} & 90.18  & 88.14  & 66.77  & 89.89  & 77.83  & 29.76  & 66.85  & 60.93  & 70.58  & 70.54  & 62.02  & 70.32  \\
    \textbf{Tip+D\&D} & \textbf{91.13} & \textbf{88.36} & \textbf{68.91} & \textbf{90.17} & \textbf{78.46} & \textbf{30.50} & \textbf{68.81} & \textbf{62.16} & \textbf{72.40} & \textbf{72.47} & \textbf{62.74} & \textbf{71.46} \\
    \bottomrule
    \end{tabular}%
    }
  \label{tab:few_shot_all}%
\end{table*}%

\begin{table*}[htbp]
  \centering
  \caption{Full results on the Cross-Domain Benchmark with RN-50 backbone. We report top-1 accuracy and “Average" is calculated by taking the mean accuracy across all ten datasets.}
    \begin{tabular}{l|cccccccccc|c}
    \toprule
          & \begin{sideways}Caltech \end{sideways} & \begin{sideways}Pets \end{sideways} & \begin{sideways}Cars   \end{sideways} & \begin{sideways}Flowers\end{sideways} & \begin{sideways}Food101\end{sideways} & \begin{sideways}Aircraft  \end{sideways} & \begin{sideways}SUN397 \end{sideways} & \begin{sideways}DTD\end{sideways} & \begin{sideways}EuroSAT\end{sideways} & \begin{sideways}UCF101\end{sideways} & \textit{Average} \\
    \midrule
    CLIP \cite{radford2021learning} & 85.88 & 83.57 & 55.70 & 61.75 & 73.97 & 15.66 & 58.8 & 40.37 & 23.69 & 58.84 & 55.82 \\
    CLIP + TPT \cite{shu2022test} & 87.02 & 84.49 & 58.46 & 62.69 & 74.88 & 17.58 & 61.46 & 40.84 & 28.33 & 60.82 & 57.66 \\
    CALIP \cite{guo2023calip} & 87.71 & 86.21 & 56.27 & 66.38 & 77.42 & 17.76 & 58.59 & 42.39 & 38.90 & 61.72 & 59.34 \\
    DiffTPT \cite{feng2023diverse} & 86.89 & 83.40 & \textbf{60.71} & 63.53 & \textbf{79.21} & 17.60 & 62.72 & 40.72 & 41.04 & 62.67 & 59.85 \\
    CuPL \cite{pratt2023does}& 89.29 & 84.84 & 57.28 & 65.44 & 76.94 & \textbf{19.59} & 62.55 & \textbf{48.64} & 38.38 & 58.97 & 60.19 \\
    TDA \cite{karmanov2024efficient} & \textbf{89.70} & 86.18 & 57.78 & \textbf{68.74} & 77.75 & 17.61 & 62.53 & 43.74 & 42.11 & 64.18 & 61.03 \\
    \midrule
    \textbf{TDA+D\&D} & 88.76 & \textbf{87.60}  & 59.92 & 68.09 & 78.18 & 18.54 & \textbf{63.05} & 44.39 & \textbf{48.37} & \textbf{65.61} & \textbf{62.25} \\
    \bottomrule
    \end{tabular}%
  \label{tab:tta_rn50}%
\end{table*}%

\end{document}